\documentclass{llncs}
\usepackage{amssymb,amsmath,subcaption}
\usepackage{makeidx}  % allows for indexgeneration
\usepackage{algorithm}
\usepackage{algorithmic}
\usepackage{epstopdf}
\newcommand{\remove}[1]{}

\usepackage{amssymb}
\usepackage{graphicx}
\usepackage{color}
\newcommand{\xhdr}[1]{\vspace{1mm}\noindent{{\bf #1.}}}

\usepackage{hyperref}
\usepackage{array}
\newcolumntype{L}{>{\centering\arraybackslash}m{3cm}}
\usepackage{wrapfig}
\usepackage{paralist}

\begin{document}
\pagestyle{headings}  % switches on printing of running heads

\mainmatter              % start of the contributions
\title{Ballpark Learning: \\ Estimating Labels from Rough Group Comparisons 
%\dnote{rename. stereotypes? educated guess? ballpark? guesstimate?}
}
\titlerunning{Ballpark Learning}  % abbreviated title (for running head)
\author{Tom Hope \and Dafna Shahaf
}
\authorrunning{Hope Shahaf} % abbreviated author list (for running head)
\tocauthor{Tom Hope, Dafna Shahaf}
\institute{The Hebrew University of Jerusalem\\
\email{tom.hope@mail.huji.ac.il, dshahaf@cs.huji.ac.il}\\ 
}

\maketitle              % typeset the title of the contribution

\begin{abstract}
We are interested in estimating individual labels given only coarse, aggregated signal over the data points. In our setting, we receive sets (``bags'') of unlabeled instances with constraints on label proportions. We relax the unrealistic assumption of known label proportions, made in previous work; instead, we assume only to have upper and lower bounds, and constraints on bag differences.
We motivate the problem, propose an intuitive formulation and algorithm, and apply our methods to real-world scenarios. Across several domains, we show how using only proportion constraints and no labeled examples, we can achieve surprisingly high accuracy. In particular, we demonstrate how to predict income level using rough stereotypes and how to perform sentiment analysis using very little information. We also apply our method to guide exploratory analysis, recovering geographical differences in twitter dialect. %Our results agree with previous findings and suggest potential new ones.
%\keywords{computational geometry, graph theory, Hamilton cycles}
\end{abstract}

\section{Introduction}
\label{sec:intro}
% \begin{enumerate}
% \item Labeling is expensive, restricts application
% \item (Patents? sometimes only know coarse signal. financial example?)
% \item like proportions
% \item what's wrong with current solutions/formulations (not reasonable to assume we know proportions, what else?)
% \item we show how to solve without knowing proportions
% \item (works on synthetic data)
% \item Two real-world applications: labeled features (show how we can classify with just one word) and the exploratory setting
% \end{enumerate}

In many classification problems, labeled \remove{problem}instances are often difficult, expensive, or time-consuming
to obtain. Unlabeled instances, on the other hand, are easier to obtain, but it is harder to use them for classification.
Semi-supervised learning \cite{chapelle2006semi} addresses this problem, using unlabeled instances together with a small amount of labeled instances to improve performance. 

We are interested in a learning setting where few, if any, labeled instances exist. Instead, we only know some coarse, aggregated signal over the data points. In particular, our instances are divided into sets (or \emph{bags}), and we are given some aggregate information about the bags; for example, we might know that one bag has a higher percentage of positive-label instances than another.

%For example, suppose we want to perform sentiment analysis on financial news, but we have no labeled articles. Instead, we have proxy indicators for market-wide sentiment, such as stock prices and trading volumes. We might hypothesize that articles tend to be more positive around days where the proxy indicators are higher; in other words, we might have class \emph{proportion} constraints.

%\begin{figure}[b!]
%\centering
\begin{wrapfigure}[12]{R}{0.5\linewidth}
\includegraphics[width=1\linewidth]{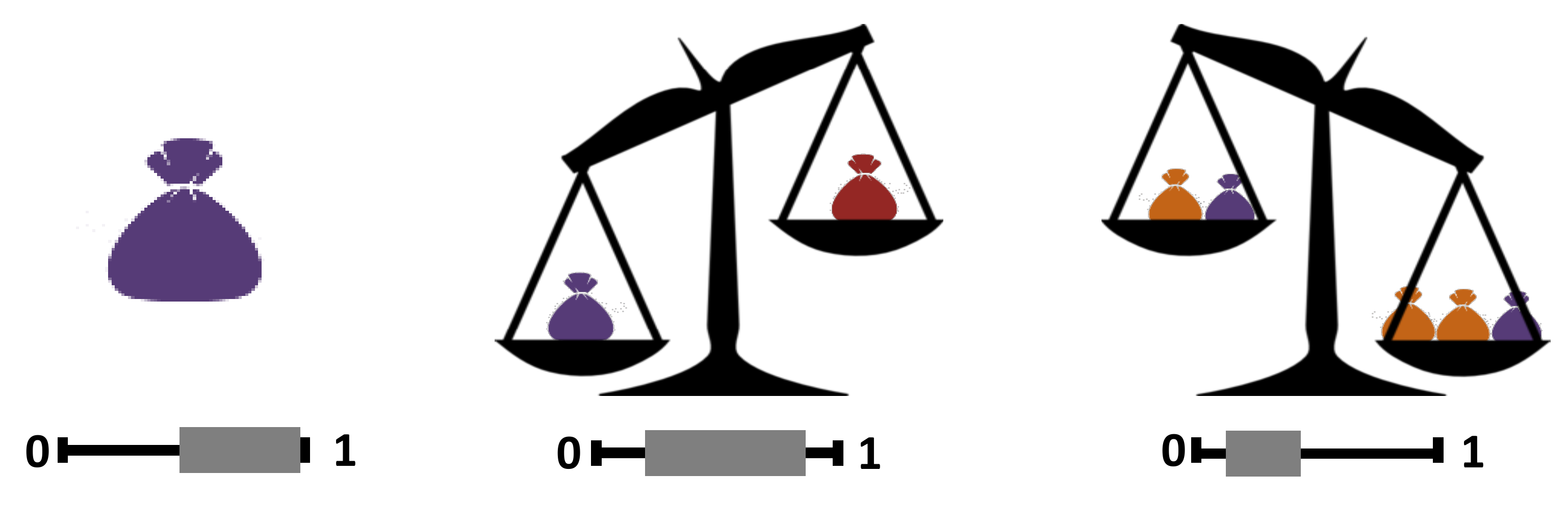}
\caption{\small We are given bags of instances and rough estimates about label proportions and differences between bags. Here, the purple bag has at least $50\%$ positive instances, more than the red bag (but the magnitude of the difference is uncertain).}
\label{fig:scale}
\end{wrapfigure}
%\end{figure}

There is recent interest in the task of estimating the labels of \emph{individual} instances given aggregate information, due to the many real-world scenarios in which such information is available. In particular, aggregate information (e.g., summary statistics) is often published for sensitive data, when one cannot publish individual statistics. Being able to estimate individual labels from such data has important implications regarding privacy and data anonymization. 

Constraining class proportions of unlabeled data has been shown to be useful for semi-supervised learning \cite{quadrianto2009estimating,Felix13svm,wager2015clustering}. 
Under this setting, we are given sets of unlabeled instances with known label proportions (for example, one bag has 30\% positive instances and 70\% negative instances). %These proportions are either obtained as an input or estimated. 
%
%%%%%%%%%%%%%%%%%%

We believe that the assumption of known proportions is unrealistic, and limits the applicability of such methods.
For example, suppose we want to classify Twitter users by political orientation. We have some information about the users (for example, the text of their tweets), but no explicit political affiliation to use as labels. We could, however, use the commonly-known fact that political orientation is correlated with geographic location. 
Thus, we can construct bags of users based on their geographic location: bags would correspond to states whose residents predominantly vote for the Republican Party (red states) or Democratic Party (blue states).

Estimating the proportion of Democrats on Twitter is hard, even using location information. Previous election data or polls are unlikely to accurately reflect the behavior of Twitter users. 
Instead of assuming known proportions, we propose a setup where our input is much weaker: we only know some constraints on bag proportions and on differences between bags. %\tnote{Polls, for example, could serves as a \textit{rough} guide as to the actual underlying proportions, rather than be considered accurate estimates.}
In other words, users from red states do not necessarily vote for the Republicans, but it is safe to expect to see more Republicans in the red-state bags. It is also reasonable %\dnote{conservative! pun intended} 
to assume that, say, at least $10\%$ of Blue-state users are Democrats. Using only this type of weak, ``ballpark'' estimates, we would like to be able to classify individual users. 

%\dnote{is this fair as an example? perhaps give census data with education, gender}

%For example, we can assume that on days where the market indicators are extremely high, the proportion of positive-sentiment articles is higher than on the market's worst days.

Figure \ref{fig:scale} demonstrates this idea. Our input includes  approximate information on label proportions in some bags (left) and pairwise comparisons between bags (middle) or sets of bags (right).
Our contributions are as follows:
%\dnote{do we even talk about sets of bags? I mean, it's a trivial extension, but still} 

%For example, we might have two bags of articles -- one from days when market indicators were high and one from when they were low -- and a constraint specifying that the percentage of positive labels needs to be bigger in the former bag.

%In this paper, we formalize the model and propose an alternating optimization algorithm. We give practical examples motivating this setup, applying our method to real-world applications: sentiment analysis of movie reviews from a very coarse signal and prediction of income from stereotypes.
%Our algorithms are designed for scenarios in which human labeling resources are scarce.
%Finally, we show how the algorithm can guide exploratory classifications, recovering geographical differences in twitter dialect.

%\tnote{First, we classify movie reviews without labeled examples. We use information such that texts containing the word ``great" are more likely to be positive than negative, and that reviews with the word ``nice" in them are more likely to be positive than those with ``terrible". We incorporate rough estimates for proportions, which could come from a sample or from domain knowledge (e.g., at least $10\%$ of the reviews that have ``great'' in them are positive).}

\begin{compactitem}
	\item  We extend the Learning from Labeled Proportions setting by proposing a new, more realistic scenario in which label proportions in each bag are not assumed to be known, but rather some constraints on them. We suggest various domains that lend themselves to this setting. 
	\item We propose a simple and intuitive bi-convex
 problem formulation and an efficient algorithm, including a novel form of cross-validation.
	\item We apply our algorithm to real data,   
 perform sentiment analysis of movie reviews from a very coarse signal, and predict income using stereotypes.
    \item We demonstrate the use of our method for exploratory analysis. We find vernacular difference in geo-tagged tweets by incorporating expressive constraints such as ``Alabama $>$ Florida $>$ New York".    
\item Our algorithm is designed to use when human labeling resources are scarce. Despite the simplicity of our methods, we achieve high accuracy with a very modest amount of input, and considerably loose (or misspecified) constraints.
      
\end{compactitem}

\section{Problem Formulation}
\label{sec:formulation}

We begin by formalizing our setting and problem.
Consider a set of $N$ training instances $\mathcal{X}_N = \{ \mathbf{x}_1, \mathbf{x}_2,\ldots, \mathbf{x}_N \}$. Each $\mathbf{x}_i$ has a corresponding \textit{unknown} label $y^{*}_i \in \{-1,1\}$. In addition, we could be given a (possibly empty) set of $L$ labeled training instances  $\mathcal{X}_L = \{ \mathbf{x}_{N+1}, \mathbf{x}_{N+2},\ldots, \mathbf{x}_{N+L} \}$ with known binary labels  $y_i$, where typically  the vast majority of our instances are unlabeled: $N \gg L$.
%
%Using notation similar to \cite{Felix13svm}, 
In addition, we are given a set of $K$ subsets of $\mathcal{X}$, which we call \emph{bags}:
$$ \mathcal{B} = \{ \mathcal{B}_1, \mathcal{B}_2, \ldots \mathcal{B}_K\} , \mathcal{B}_k \subseteq \mathcal{X}_N \cup \mathcal{X}_L.$$
Note that bags $\mathcal{B}$ may overlap, and do not have to cover all training instances $\mathcal{X}_N$. Let $p_k$ be the proportion of positive-labeled instances in bag $\mathcal{B}_k$:
\begin{equation}
\label{eq:p_def}
p_k = |\{i:i \in \mathcal{B}_k, y^{*}_i  =1\}/ {|\mathcal{B}_k|}
\end{equation}
(where $y^{*}_i$ is replaced with $y_i$ for instances $\mathbf{x}_i \in \mathcal{X}_L$).  Previous work \cite{quadrianto2009estimating} tackled the case of \emph{known} label proportions, suggesting that precise proportions could be estimated using sampling. However, obtaining accurate estimates could be costly or impractical (e.g., for bags with high label skew). 
In this work we {do not assume to know} $p_k$. Rather, we are given weaker prior knowledge, in the form of constraints on proportions. We allow constraints of the following forms:
\begin{compactitem}
	\item \textbf{Lower and upper bounds} on bag proportions:  $l_k \leq p_k \leq u_k$
	\item \textbf{Bag difference bounds}:  $ 0 \leq l_{k_{12}} \leq p_{k_1} - p_{k_2} \leq u_{k_{12}}$ 
\end{compactitem}
We are especially interested in the case where very little information is known: constraints are \emph{loose}, and specified only for a small subset of the bags.

Our goal is to predict a label for each $\mathbf{x}_i$, using a function $f(\mathbf{x}) = \text{sign}(\mathbf{w}^{T}\varphi(\mathbf{x}))$, where $\mathbf{w}$ is a weight vector and  $\varphi(\cdot)$ is a feature map (to simplify notation we drop a bias term $\mathbf{b}$ by assuming a vector $\mathbf{1}_{N+L}$ is appended to the features). To attain the classification goal, we use a maximum-margin approach.  Let $\mathcal{R}$ be the subset of $\mathcal{B}$ for which we have upper and/or lower bounds. Let $\mathcal{D}$ be the set of tuples $(\mathcal{B}_{k_1},\mathcal{B}_{k_2})$ for which we have difference bounds. To solve this problem we directly model the latent variable $\mathbf{y}^{*}$ -- the vector of unknown labels $y^{*}_i \in \{-1,1\}$, in an alternating optimization approach.

Noting that (\ref{eq:p_def}) can be written as $p_k  =  \frac{\sum_{i \in \mathcal{B}_{k}} y^{*}_i}{2|\mathcal{B}_{k}|} + \frac{1}{2}$, we formulate the following bi-convex optimization problem: 
\begin{equation}
\label{eq:prob_biconv}
\begin{aligned}
\underset{\mathbf{y,w,\xi}}{\text{argmin}} & \frac{1}{2}  \mathbf{w}^T\mathbf{w} +  \frac{C}{N} \sum\limits_{i=1}^{N} \text{max}(0,1 - y_i\mathbf{w}^{T}\varphi(\mathbf{x}_i)) + \frac{C_L}{L}\sum\limits_{j=N+1}^{N+L} \xi_j \\
 s.t. 
& -1 \leq y_i \leq 1 \quad  \forall i \in 1,\ldots,N \\
&  y_j\mathbf{w}^{T}\varphi(\mathbf{x}_j) \geq 1-\xi_j \quad \forall j \in \{N+1,\ldots, N+L\} \\
& \xi_j \geq 0 \quad \forall j \\
& l_k \leq  \hat{p}_k \leq u_k \quad  \forall \{k : \mathcal{B}_{k} \in \mathcal{R} \} \\
& l_{k_{12}} \leq \hat{p}_{k_1}-\hat{p}_{k_2} \leq u_{k_{12}} \quad
\forall \{k_1 \neq k_2 : (\mathcal{B}_{k_1},\mathcal{B}_{k_2}) \in \mathcal{D} \},
\end{aligned}
\end{equation}

where $\hat{p}_k = \frac{1}{2|\mathcal{B}_{k}|}\sum\limits_{i \in B_k} y_i + \frac{1}{2}$ is the estimated positive label proportion in bag $\mathcal{B}_{k}$, $l_k$ (or $u_k$) can be $0$  ($1$) if not given as input, and analogously  for difference bounds $l_{k_{12}} (u_{k_{12}})$. $C$ and $C_L$ are cost hyperparameters for unlabeled and labeled instances, respectively.
Intuitively, the second term in the objective function  helps find a weight vector $\mathbf{w}$ accurately predicting $\mathbf{y}$, and constraints ensure that we find an assignment to $\mathbf{y}$ that satisfies proportions constraints. $C_L$ controls how much weight we give to our labeled instances versus our prior knowledge on $\mathcal{B}$. In our experiments we do not use any labeled instances, thus $C_L = 0$. 

\section{Algorithm}

%\dnote{rephrase}
%We have formalized our problem as a non-convex optimization problem.  In this section we use a simple relaxation of the label space, obtain a bi-convex program and propose an intuitive alternating optimization algorithm to solve it.

We have formalized our problem as a bi-convex optimization problem -- holding either $\mathbf{w}$ or $\mathbf{y}$ fixed, we get a convex problem. We thus propose the following intuitive alternating algorithm to solve it.
\begin{itemize}
	\item For a fixed $\mathbf{w}$, solve for $\mathbf{y}$:
	\begin{equation}
	\label{eq:solve_y}
	\begin{aligned}
	\underset{\mathbf{y}}{\text{argmin}} & \frac{1}{N} \sum\limits_{i=1}^{N} \text{max}(0,1 - y_i\mathbf{w}^{T}\varphi(\mathbf{x}_i))\\
	 s.t. 
	& -1 \leq y_i \leq 1 \quad  \forall i \in 1,\ldots,N \\
	& l_k \leq  \hat{p}_k \leq u_k \quad  \forall \{k : \mathcal{B}_{k} \in \mathcal{R} \} \\
	& l_{k_{12}} \leq \hat{p}_{k_1}-\hat{p}_{k_2} \leq u_{k_{12}} \quad
	\forall \{k_1 \neq k_2 : (\mathcal{B}_{k_1},\mathcal{B}_{k_2}) \in \mathcal{D} \},
	\end{aligned}
	\end{equation}
	\item For a fixed $\mathbf{y}$,  solve w.r.t  $\mathbf{w}$:  
	\begin{equation}
	\label{eq:solve_w}
	\begin{aligned}
	\underset{\mathbf{w}}{\text{argmin}} & \frac{1}{2}  \mathbf{w}^T\mathbf{w} + \frac{C}{N} \sum\limits_{i=1}^{N} \text{max}(0,1 - y_i\mathbf{w}^{T}\varphi(\mathbf{x}_i)) + \frac{C_L}{L}\sum\limits_{j=N+1}^{N+L} \xi_j\\
	 s.t. \ \ 
	&  y_j\mathbf{w}^{T}\varphi(\mathbf{x}_j) \geq 1-\xi_j \quad \forall j \in \{N+1,\ldots, N+L\} \\
	& \xi_j \geq 0 \quad \forall j \\
	\end{aligned}
	\end{equation}
\end{itemize}
Intuitively, the first step finds an assignment to $\mathbf{y}$ that is ``close" to predictions made by applying weights $\mathbf{w}$, and also satisfies proportions constraints. The second step re-adjusts $\mathbf{w}$. Our alternating algorithm for this bi-convex problem is thus guaranteed to descend, decreasing the objective in every iteration.

In practice, we replace $\mathbf{y}$ with $Sign(\mathbf{y})$ ($Sign(\cdot)$ applied elementwise) in order to use efficient off-the-shelf SVM solvers (See Figure \ref{alg:Alter}.
). Empirically, in most cases
we observed that $\mathbf{y}$ were very close to either $1$ or $-1$. 
%\dnote{think about competition / special cases}

To start off the alternation, we need to initialize $\mathbf{w}$. Specific label proportions constraints are handled by modeling the latent $\mathbf{y}^*$ directly, which is only possible in our alternating scheme once a vector $\mathbf{w}$ is fixed. Thus, we start the alternating optimization process by first solving the following simple convex program, which uses only the partial order between bags. Let the set of pairwise orderings $\mathcal{P}$ be the set of all tuples $(\mathcal{B}_{k_1},\mathcal{B}_{k_2})$ such that $p_{k_1} \geq p_{k_2}$. To find our initial $\mathbf{w}$ we solve: 
\begin{equation}
\begin{aligned}
& \underset{\mathbf{w},\mathbf{\xi}}{\text{argmin}} \frac{1}{2}  \mathbf{w}^T\mathbf{w} + \frac{1}{|\mathcal{P}|}\sum\limits_{p=1}^{|\mathcal{P}|} \xi_p + \frac{C_L}{L}\sum\limits_{j=N+1}^{N+L} \xi_j \\
 s.t. 
& \quad y_j\mathbf{w}^{T}\varphi(\mathbf{x}_j) \geq 1-\xi_j \quad \forall j \in \{N+1,\ldots, N+L\} \\
& \mathbf{w}^{T}\frac{1}{|\mathcal{B}_{k_1}|}\sum\limits_{i \in \mathcal{B}_{k_1}} \varphi(\mathbf{x}_i) \geq \mathbf{w}^{T}\frac{1}{|\mathcal{B}_{k_2}|}\sum\limits_{i \in \mathcal{B}_{k_2}} \varphi(\mathbf{x}_i) -\xi_p \\
& \quad \forall \{k_1 \neq k_2 : (\mathcal{B}_{k_1},\mathcal{B}_{k_2}) \in \mathcal{P} \},
\end{aligned}
\label{eq:prob1}
\end{equation}
 
\begin{wrapfigure}[19]{r}{0.5\textwidth}
%\begin{minipage}
%\begin{algorithm}
\fbox{\parbox{0.5\textwidth}{
{\bf Input:} $\mathbf{x}, \mathcal{R}, \mathcal{D},C$ \\
	\begin{enumerate}
		\item \textbf{Init} $\mathbf{w}^0$: $\mathbf{w}^0 \leftarrow$ Solution to (\ref{eq:prob1}) \\
		\item \textbf{Repeat}
		\begin{enumerate}
			\item Solve (\ref{eq:solve_y}) for $\mathbf{y}^t$ w.r.t $\mathbf{w}^{t-1}$
			\item Solve an SVM problem for $\mathbf{w}^t$ w.r.t $Sign(\mathbf{y}^t)$ and cost parameter $C$
		\end{enumerate}
		\textbf{until} $\frac{||\mathbf{w}^{t} -\mathbf{w}^{t-1}||^{2}_{2}}{||\mathbf{w}^{t-1}||^{2}_{2}} \leq 10^{-5}$
	\end{enumerate}
	\textbf{Return} $\mathbf{w}$ %corresponding to lowest SVM objective function over $\mathcal{C}_{\text{grid}}$
    }}
	\protect\caption{Alternating Algorithm}
	\label{alg:Alter}	
%\end{algorithm}
%\end{minipage}
\end{wrapfigure}

The second constraint in Problem \ref{eq:prob1} amounts to representing bags with their ($\mathbf{w}$-weighted) mean in feature-space. Note that in order for a bag  $\mathcal{B}_{k}$ to be well-approximated by its mean in feature-space, $\mathcal{B}_{k}$ should induce a low-variance distribution over bag instances. This is a strong assumption, but yields a simple quadratic program easy to solve quickly with standard solvers, and empirically leads to good starting points in parameter-space. We additionally note that when $C_L = 0$ (no labels), we recover as a special case the Multiple-Instance (MI) ranking problem proposed in the image-retrieval framework of \cite{Hu08ranking}, albeit with a different objective (we are interested in classifying instances rather than learning to rank bags).
We note that in Problem \ref{eq:solve_y}, we impose hard constraints on label proportions. Certain sets of constraints could, of course, be infeasible. In this case, a practitioner might adjust the constraints, or simply make them soft (by adding slack variables).

\xhdr{Optimizing $C$}
\label{sec:optimC}
In practice, we need to tune hyperparameter $C$. This is typically done with cross-validation (CV) grid search, measuring performance on held-out data. However, standard CV is impossible here, as we have no labeled examples. 

We thus develop a novel variant of CV, suited for our setting. We run $K$-fold CV, splitting each bag $\mathcal{B}_{k}$ into training and held-out subsets. The intuition is that the label proportion in uniformly-sampled subsets of a bag is similar to the proportion $p_k$ in the entire bag.
For each split we run Algorithm \ref{alg:Alter} on training bags, and then compute by how much constraints are violated on \textit{held-out} bags. More formally, we compute the average 
deviations from bounds, $max(\hat{p_k}-u_k,0)$, $max(l_k-\hat{p_k},0)$ for $\hat{p_k}$ the estimated label proportion in the held-out subset of bag $k$. We do so over a grid, and select the $C$ with lowest average violation.

%More formally, denote by $\hat{p_k}$ the estimate for $p_k$ on a test-split in a CV iteration. For each constraint in sets $\mathcal{R}, \mathcal{D}$, we compute $max(\hat{p_k}-u_k,0)$, $max(l_k-\hat{p_k},0)$, and so on for all relevant constraints.  We obtain the average of these deviations for each $C$, and select the $C$ with lowest average deviation. Ties are broken by taking lower values (less ``rigid" penalties). Of course, violations could be measured for only a subset of $\mathcal{R}, \mathcal{D}$. In practice, we find in our experiments that checking for violations of upper and lower bounds on individual $p_k$ suffices.

\section{Evaluation}
\label{sec:app}

%We now demonstrate several applications of our model:  in classification with a weak signal, and exploratory analysis. Obtaining labeled data is often an expensive procedure, and only few labeled instances are available at training time. This problem has spawned much research over the years, including in the fields of semi-supervised, weakly-supervised, unsupervised and active learning. In all of our applications presented below, we thus assume no labels at all are given during training time. However, as shown above, our method is directly geared to receiving some training labels, too, which could help guide the learning process. 

In order to evaluate our algorithm, we prepared the following datasets:

\begin{itemize}
	\item {\bf Movie Reviews:} The Movie Reviews dataset \cite{PangLee:04a}
	contains 1000 positive and 1000 negative movie reviews written before 2002. The task is to classify the sentiment of movie reviews as positive or negative. 
	\item {\bf Census:} The Adult dataset \cite{adult} ($48842$ instances) is from the Census bureau. The task is to predict whether a given adult makes more than \$50,000 a year based on attributes such as education, hours of
	work per week, etc.
\end{itemize} 

For each of the classification tasks described, we run $10$-fold cross-validation and report average results (note that labels are used only for testing).
For text classification tasks, feature map $\varphi(\cdot)$ is the standard TF-IDF features.

We formed bags corresponding to the different tasks (see below), demonstrating the wide applicability of the setting and our approach. In order to test our method's robustness we used approximate constraints, at times violating the true underlying proportions.

\xhdr{Baselines}
To the best of our knowledge, no other method aims to solve the problem of Section \ref{sec:formulation}. Thus, we compare ourselves to three natural baselines.
\begin{compactitem}
	\item{\bf ``High vs. low":} One reasonable approach in our setting is to create two sets of instances: The ``high'' set contains instances from bags with the highest label proportions, and the ``low'' set -- from bags with the lowest proportions. The idea is to pretend all instances in the ``high'' set are positive, and in the ``low'' set -- negative, and learn a classifier with the noisy labels. To make the baseline stronger, we use grid search to optimize hyper-parameter $C$ (chosen from a commonly used grid for SVM $C$ values, $[10^{-4},10^{-3},\ldots,10^3,10^4]$, with $10$-fold cross-validation and selecting $C$ with best average). To counter the class-imbalance created, we apply a weighted SVM.
\item{\bf Supervised SVM:} Our method does not need labeled instances, but instead uses weaker, aggregate information. To show how many labels are needed to obtain comparable results to our method, 
	we report SVM results over a labeled training set (note that this information is not available to our algorithm). We use grid search to optimize hyper-parameter $C$ as above.
    
    \item{\bf Learning from labeled proportions} For the census data set, we compare our method's performance to results reported in \cite{quadrianto2009estimating} using known label proportions with various algorithms. Note that our method does not have access to the exact label proportions.
\end{compactitem} 

For our method, we select $C$ using the constraint-violation approach described in the previous section. %The small values of the  grid help prevent our alternating algorithm from ``locking-in" too quickly on sub-optimal solutions by deliberately imposing low penalties on errors early on. This allows more flexibility in the search of label-space and parameter-space. 

We run the procedure for a maximum of $200$ iterations, with convergence typically occurring long before. A typical iteration (for one value of $C$, one CV split) took at most a few seconds on a standard laptop. Our data is available on  \url{https://github.com/ttthhh/ballpark.git}.

\subsection{One-Word Classifier}
Our first task is to classify sentiment of movie reviews. Our
goal is not to compete with the host of previous
sentiment-analysis algorithms \cite{pang2008opinion} in terms of accuracy, but rather
to provide a light-weight tool when very little information
and resources are available: a ``poor-man's" classifier. In this section, we show how we are able to obtain good
results while assuming very scarce prior knowledge with simple, clean tools.
 
We envision a practitioner who knows a very simple fact -- that reviews containing the word ``great" are more likely to be positive than negative, but far from exclusively: many positive reviews do not use the word ``great", and some negative reviews do use it (``horrific performance by a usually great actor").

We construct three bags: $\mathcal{B}_\text{great},\mathcal{B}_\text{good},\mathcal{B}_\text{bad}$, each containing reviews with the corresponding word in them (note the bags are not necessarily disjoint).
For the three bags created on training set instances ($10$-fold CV) we find that $|\mathcal{B}_\text{great}|  \approx 700$, $|\mathcal{B}_\text{good}| \approx 630$, $|\mathcal{B}_\text{bad}| \approx 160$, $p_\text{great} \approx 0.6 $,$p_\text{good} \approx 0.45 $, $p_\text{bad} \approx 0.25$. 

For simplicity, we assume no labels are given, but the practitioner has rough estimates for proportions. This information could come from a sample or from domain knowledge. 
In our experiment, we assume an upper bound on the bag with the highest proportion and a lower bound for each bag. 
We used a weak bound for each bag, underestimating it by $50\%$. We also assumed that $p_\text{great} > p_\text{good} > p_\text{bad}$. Again we use a weak bound, overestimating the real difference by $33\%$. In Section \ref{sec:analysis} we explore how the tightness of the constraints affects accuracy, showing our method is robust to loose constraints.

%\begin{itemize}
%	\item \textbf{Upper bounds on bag proportions} As above, we assume to be given an upper bound only on $p_{k_{max}}$, in this case  $p_\text{great} \leq 0.6$. 
%	\item \textbf{Lower bounds} For each true $p_k$, we take as a lower bound $0.5*p_k$, i.e. $p_\text{great} \geq 0.3, p_\text{good} \geq 0.2, p_\text{bad} \geq 0.1$
%	\item \textbf{Bag difference bounds}  We use $2$ constraints: $ p_\text{great} - p_\text{good} \geq 1.33\times0.15 (\approx0.2), p_\text{good} - p_\text{bad} \geq 1.33\times 0.2 (\approx 0.27)$.
%\end{itemize}

For the \textit{``high vs. low"} baseline, we take bag $\mathcal{B}_\text{great},\mathcal{B}_\text{good}$ as the positive class, and $\mathcal{B}_\text{bad}$ as the other.
As seen in Table \ref{table:Movie}, our method outperforms this naive baseline, and competes with supervised SVM trained on considerable amounts of labeled examples. Given fewer labels, supervised SVM is inferior to our label-free method: providing SVM with 25 labeled instances leads to accuracy of $0.51$, $50$ labels to accuracy of $0.63$, and $75$ labels increases accuracy to $0.69$. 

To test stability, we run the same experiment using different words to create the bags. The results are similar. Table \ref{table:Movie} shows the results using ``excellent'',``nice'', and ``terrible''. To make sure the classifier is not learning our input words, we test removing these words (e.g., ``good'') from the documents. In our experiment, the removal reduced accuracy by less than $1\%$.

\begin{table*}[t]
	\centering
	\caption{\small \textbf{Movie results} for different sets of bags based on different choices of words. Our method outperforms the naive SVM baseline, and rivals a supervised SVM with a considerable number of labels. }
	\label{table:Movie}
    {\small
	\begin{tabular}{|c|c|c|}
		\hline	
		{\bf Method}  &{\bf  $\mathcal{B}_\text{great},\mathcal{B}_\text{good},\mathcal{B}_\text{bad}$ } & {\bf $\mathcal{B}_\text{excellent},\mathcal{B}_\text{nice},\mathcal{B}_\text{terrible}$ }\\
		\hline
		{Bag constraints} &  \textbf{0.71} & \textbf{0.73} \\
		``high vs. low" SVM &  0.52 & 0.55 \\ \hline
		Supervised SVM & 100 labels (0.71) & 100 labels (0.71)  \\	\hline
	\end{tabular}
    }
\end{table*}

\subsection{Learning from Stereotypes}

In this section we simulate a scenario frequently occurring in practice. We have a large sample of individuals, and would like to predict their level of income using socio-demographic information. One variable that is known to be correlated with income is \emph{education level}. This information is difficult to obtain (budgetary constraints, privacy issues, respondents' reluctance etc.) and is available only for a small sub-sample. In addition, we have no labels -- individuals with known income. We do have ballpark-estimations on income proportions for different education levels, and the difference between them (based on an earlier census, expert assessments or other external sources).

In our first experiment we construct bags based on education level: $\mathcal{B}_\text{Masters}$, $\mathcal{B}_\text{Bachelors}$,$\mathcal{B}_\text{Some-college}$, $\mathcal{B}_\text{High-School}$. Over $20$-fold CV (size of training set $\approx 1220$) we find that $|\mathcal{B}_\text{Masters}|  \approx 90$, $|\mathcal{B}_\text{Bachelors}| \approx 265$, $|\mathcal{B}_\text{some-college}| \approx 360, |\mathcal{B}_\text{High-School}| \approx 520$, $p_\text{Masters} \approx 0.55 $,$p_\text{Bachelors} \approx 0.42 $, $p_\text{some-college} \approx 0.19,p_\text{High-School} \approx 0.16$.

We use similar constraints to the previous section, but remove all lower bounds on bags, thus incorporating even less prior information than before. For the \textit{SVM using ``high vs. low"} baseline, we use $\mathcal{B}_\text{Masters}$, $\mathcal{B}_{Bachelors}$ as one class, and $\mathcal{B}_\text{some-college},\mathcal{B}_\text{High-School}$ as the other. 

We start with basic features: age, gender, race. After assigning individuals to education bags, we discard education features from the data -- we assume not to have this information at test time (only for a small sub-sample available for training). We do retain those features for the \textit{Supervised SVM} baseline. Our method achieves cross-validation accuracy of $0.74$, while the baseline achieves $0.57$. Supervised SVM, even with $1000$ labeled examples, only reaches $0.71$. 

We also experiment with using less bags (removing ``Masters"), and with an expanded feature set (age, race, gender, hours-per-week, capital-gain, capital-loss).  See Table \ref{table:census} for results. Here too, our method outperforms the baseline, and rivals supervised SVM with $900$ labels.  

Of course, we are not limited to using bags based on only education level. Another well-known correlation is between gender and income. Thus, we can also slice the data into bags based on education \textit{and} gender. In another experiment we create $6$ bags, $\mathcal{B}_\text{Bachelors+Female}$, $\mathcal{B}_\text{Some-college+Female}$, $\mathcal{B}_\text{High-School+Female}$, $\mathcal{B}_\text{Bachelors+Male},\mathcal{B}_\text{Some-college+Male}, \mathcal{B}_\text{High-School+Male}$. There are stark differences in label proportions between the groups, notably in favor of males.

For the SVM using \textit{`high vs. low"} baseline, we try two different class assignments. We start from Bachelors vs. everyone else. (It could seem more natural to take, for example, $\mathcal{B}_\text{Bachelors+Male}$ as the ``high'' bag and $\mathcal{B}_\text{High-School+Female}$ as ``low'', but this results in too small a sample). The baseline performed relatively well (Table \ref{table:census})  due to good class separation. However, when we tried females vs. males, performance of our method remained stable (with highest accuracy), but the baseline suffered a drastic drop (Table \ref{table:census}). This highlights the difficulty of using this baseline when using multiple bags based on richer information: it is not immediately clear how to create two well-separated classes. On the other hand, our method naturally compares groups based on given constraints.

\xhdr{More Baselines}
Finally, we report classification accuracy on the same dataset, taken from \cite{quadrianto2009estimating}. The authors create two artificial bags, one retaining original label proportions and another containing only one class. With these bags, their method (using known proportions) achieved 0.81 accuracy. They also report results for Kernel Density Estimation (0.75), Discriminative Sorting -- a supervised method (0.77), MCMC sampling (0.81), and a baseline of predicting the major class (0.75). Our method achieves comparable performance despite having much less information on label proportions, fewer features, and using more realistic bags.

\begin{table*}[t]
	\centering
    \caption{\small \textbf{Census results} for different sets of bags. Our method outperforms the naive SVM baseline, and rivals a supervised SVM with many labeled examples. }
	\label{table:census}
    {\small 
	\begin{tabular}{|c|L|L|}
		\hline	
		{\bf Method}  & \multicolumn{1}{m{3.1cm}|}{\bf 
		Education bags}  & \multicolumn{1}{m{3.1cm}|}{\bf  Edu + Gender}\\
		\hline
		\textbf{Bag constraints} &  0.75 & 0.77 \\
		``high vs. low" SVM (Bachelors vs. other) &  0.52 & 0.6 \\
		``high vs. low" SVM (Female vs. Male) &  - & 0.38 \\
		Supervised SVM & 0.75 (900 labels) & 0.77 (900 labels) \\	\hline
	\end{tabular}
    }
\end{table*}

%\subsection{20 Newsgroups: Arbitrary Bags}

%\dnote{fit this in}
%For the three  \textbf{20 newsgroups} tasks, we use built-in splits provided in python package scikit-learn \cite{Pedregosa2011sklearn}. 
\subsection{Sensitivity Analysis \label{sec:analysis}}

In this section we give a short demonstration of how the tightness of constraints could affect model performance. We create artificial bags and vary the tightness of some constraints, reporting accuracy. This is a preliminary study, serving to illustrate some of the different factors that come into play. 

We use the {\bf 20 Newsgroups} dataset \cite{20news} containing approximately 20,000 posts across 20 different newsgroups. Some of the newsgroups are closely related (e.g., comp.sys.ibm.pc.hardware and comp.sys.mac.hardware), while others are further apart (rec.sport.hockey and sci.space). The task is to classify messages according to the newsgroup to which they were posted.     

We assume predefined bags and vary constraints on label proportions within and between bags. We do not use any labeled data at training time.

We examine three binary classification tasks, between different categories of posts: \textit{space} vs. \textit{medicine}, \textit{ibm.pc} vs. \textit{mac}, and \textit{hockey} vs. \textit{baseball}. For each of these binary classification tasks, we create six bags of training instances $\mathcal{B} = \{ \mathcal{B}_1, \mathcal{B}_2, \ldots \mathcal{B}_6\}$. The sizes of each bag are $|\mathcal{B}_1| = |\mathcal{B}_2| = 200, |\mathcal{B}_3| = |\mathcal{B}_4| = 50, |\mathcal{B}_5| = |\mathcal{B}_6| = 100$. We thus use only $650$ instances in this case -- about half of the $1187$ in the training set. The real label proportions within each bag are $p_1 = p_2 = 0.5, p_3 = p_4 = 0.3, p_5 = p_6 = 0.2$. 

We test the effects of three different types of constraints, corresponding to common types of aggregate information: 
\begin{compactitem}
	\item \textbf{Upper bounds on bag proportions:} Let $k_{max}$ be the index of the bag with the highest proportion. We assume an upper bound multiplicative factor only on this bag: $p_{k_{max}}\times u_m$, where we control factor $u_m$. 
	\item \textbf{Lower bounds:} For each true $p_k$, we take as a lower bound $l_\text{p}*p_k$.
	%, i.e. $p_1,p_2 \geq l_\text{p}\times 0.5$, $p_3,p_4 \geq l_\text{p}\times 0.3, p_5,p_6 \geq l_\text{p}\times 0.2$. 
	\item \textbf{Bag difference bounds:}  For each true $p_{k_1} - p_{k_2}$ such that $p_{k_1} \geq p_{k_2}$, we lower-bound the difference with  $l_\text{d}\times(p_{k_1} - p_{k_2})$. 
	%For example, $p_1 - p_5 \geq l_\text{d}\times0.3$.
\end{compactitem}
 \begin{wrapfigure}[41]{L}{0.45\linewidth}

 	\begin{subfigure}{0.5\textwidth}
 		\includegraphics[width=0.82\linewidth, height=4cm]{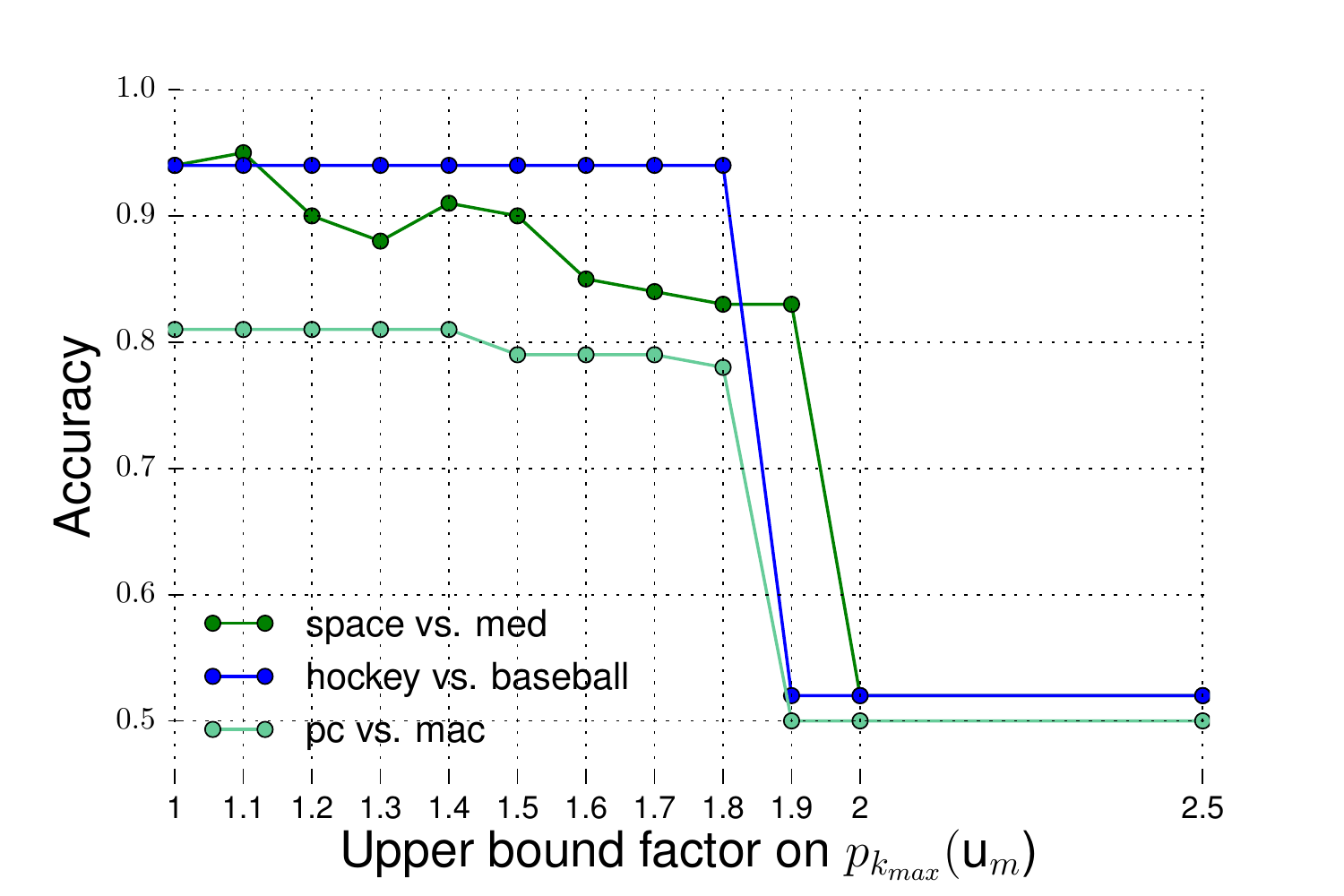} 
        
        \hspace{.35\textwidth} \small(a)
 		\label{fig:subim1}
 	\end{subfigure}
 	\begin{subfigure}{0.5\textwidth}
 		\includegraphics[width=0.82\linewidth, height=4cm]{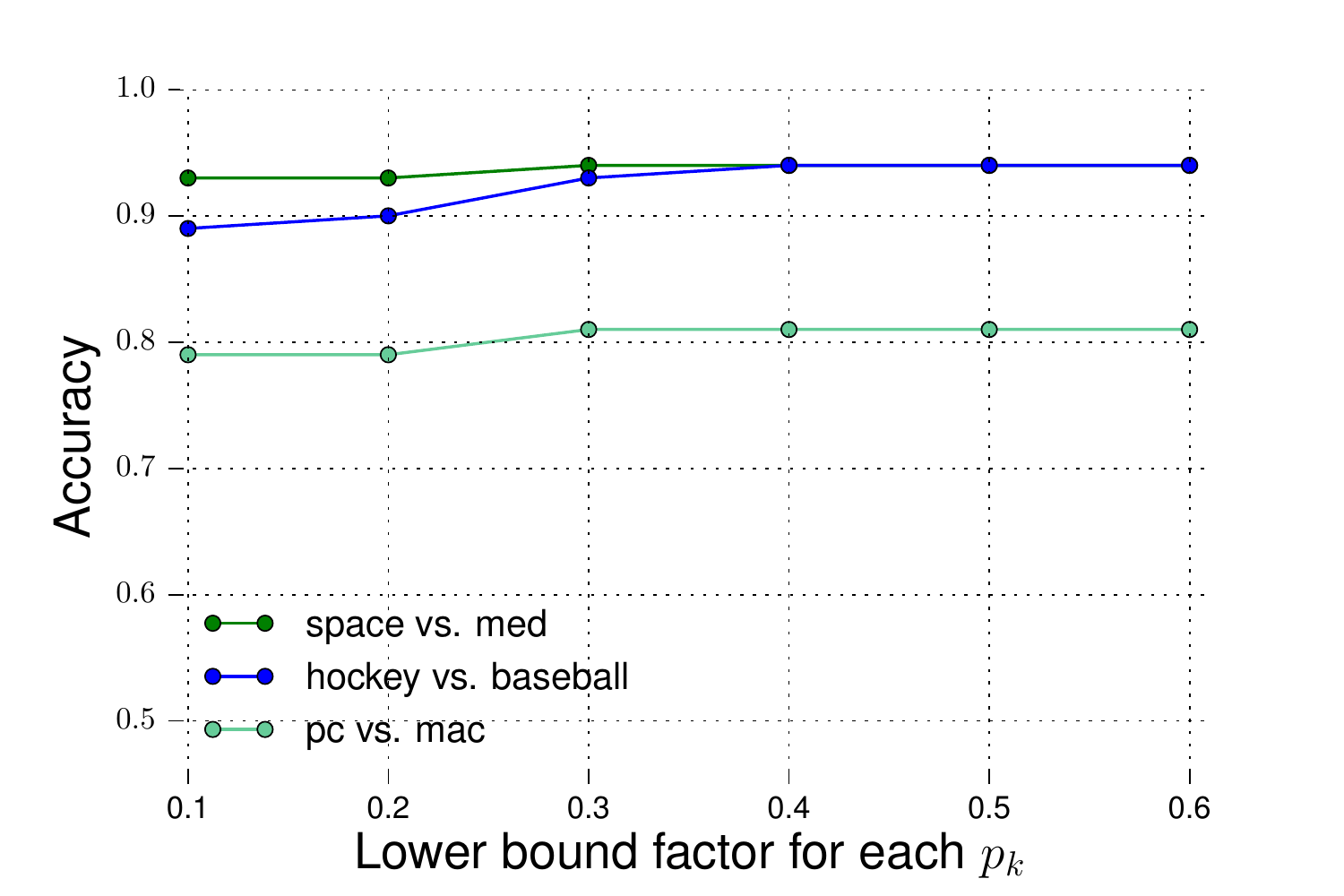} 
        
        \hspace{.35\textwidth} \small(b)
 		% \caption{Caption 2}
 		\label{fig:subim2}
 	\end{subfigure}
 	\begin{subfigure}{0.5\textwidth}
 		\includegraphics[width=0.82\linewidth, height=4cm]{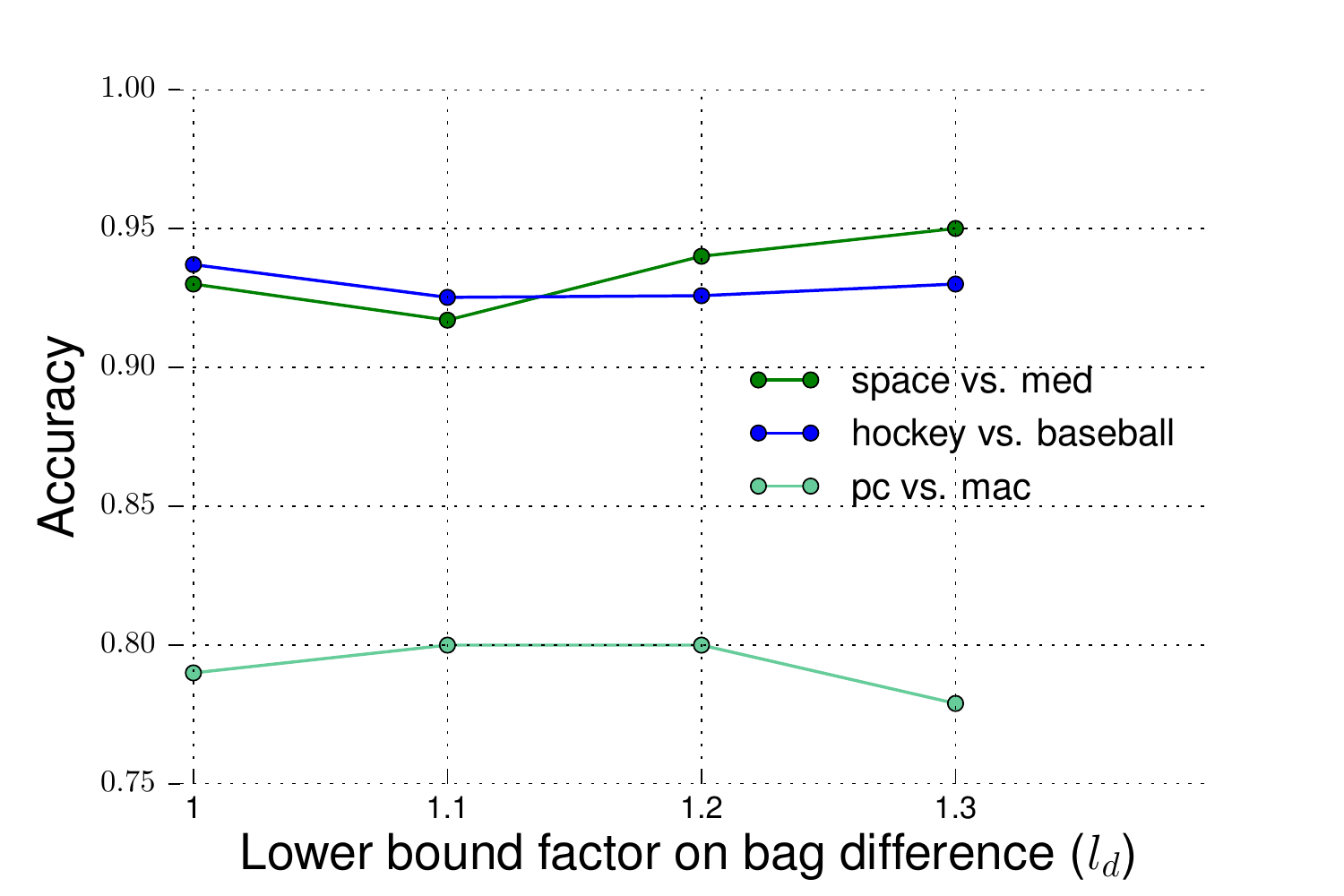} 
        
        \hspace{.35\textwidth} \small(c)
 		% \caption{Caption 2}
 		\label{fig:subim0}
 	\end{subfigure}
% \begin{subfigure}{0.5\textwidth}\hspace{.4\textwidth}(d)\\
%  			\includegraphics[width=0.9\linewidth, height=4cm]{upper_diff}
%  			% \caption{Caption 2}
%  			\label{fig:subim3}
%  		\end{subfigure}
 	\caption{\small \textbf{Constraint effects}. Accuracy results on a validation set: (a) Varying upper-bound factor on highest $p_k$  (b) Varying individual lower-bound factor (c) Varying lower bound on bag differences. Results remain fairly robust (with fluctuations due to small-sample noise). The graph stops abruptly where constraints are no longer feasible.}
 	\label{fig:20news}
 \end{wrapfigure}

\remove{ \begin{figure*}[bt]
 	\begin{subfigure}{0.5\textwidth}\hspace{.4\textwidth}(a)\\
 		\includegraphics[width=0.9\linewidth, height=4cm]{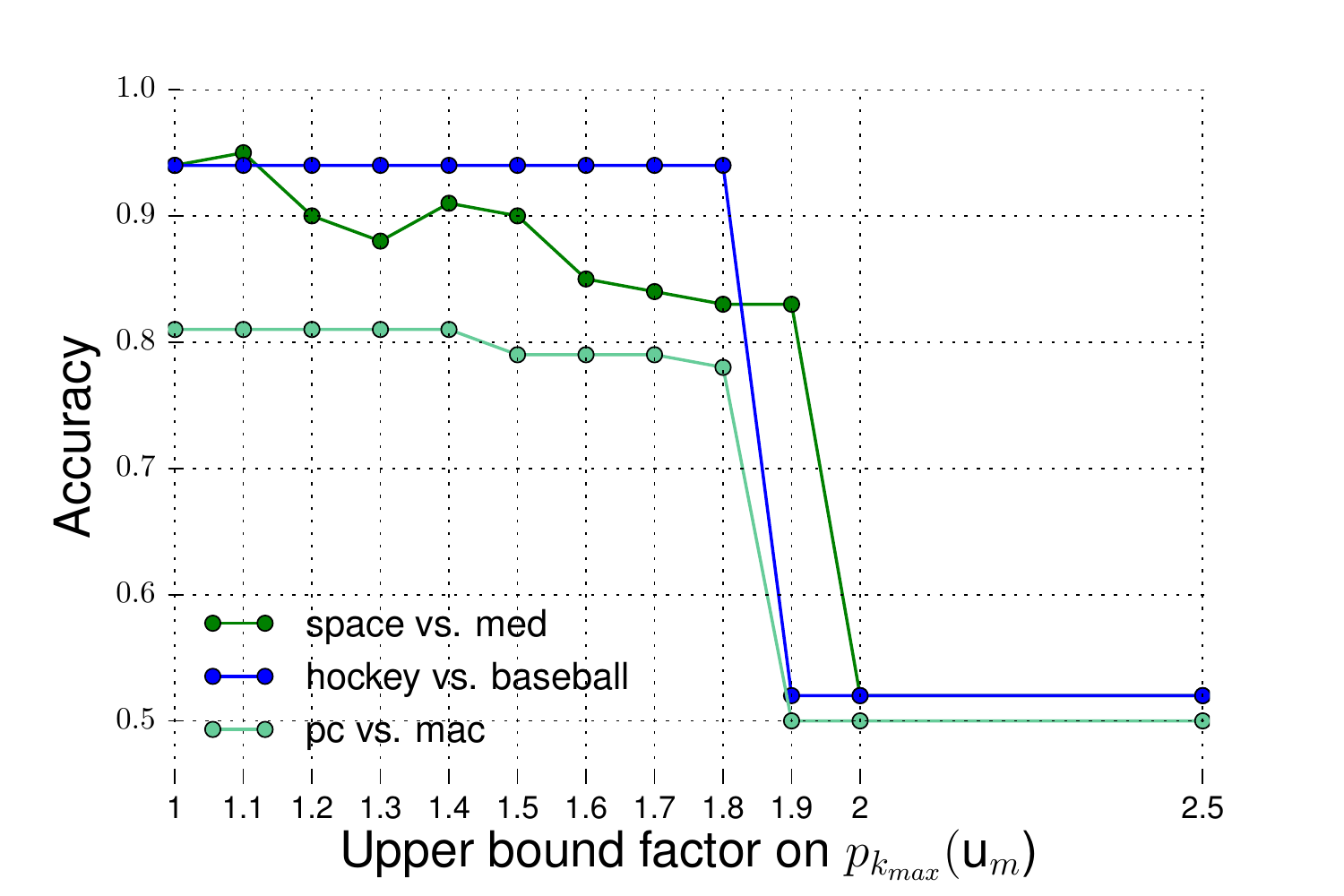} 
 		\label{fig:subim1}
 	\end{subfigure}
 	\begin{subfigure}{0.5\textwidth}\hspace{.4\textwidth}(b)\\
 		\includegraphics[width=0.9\linewidth, height=4cm]{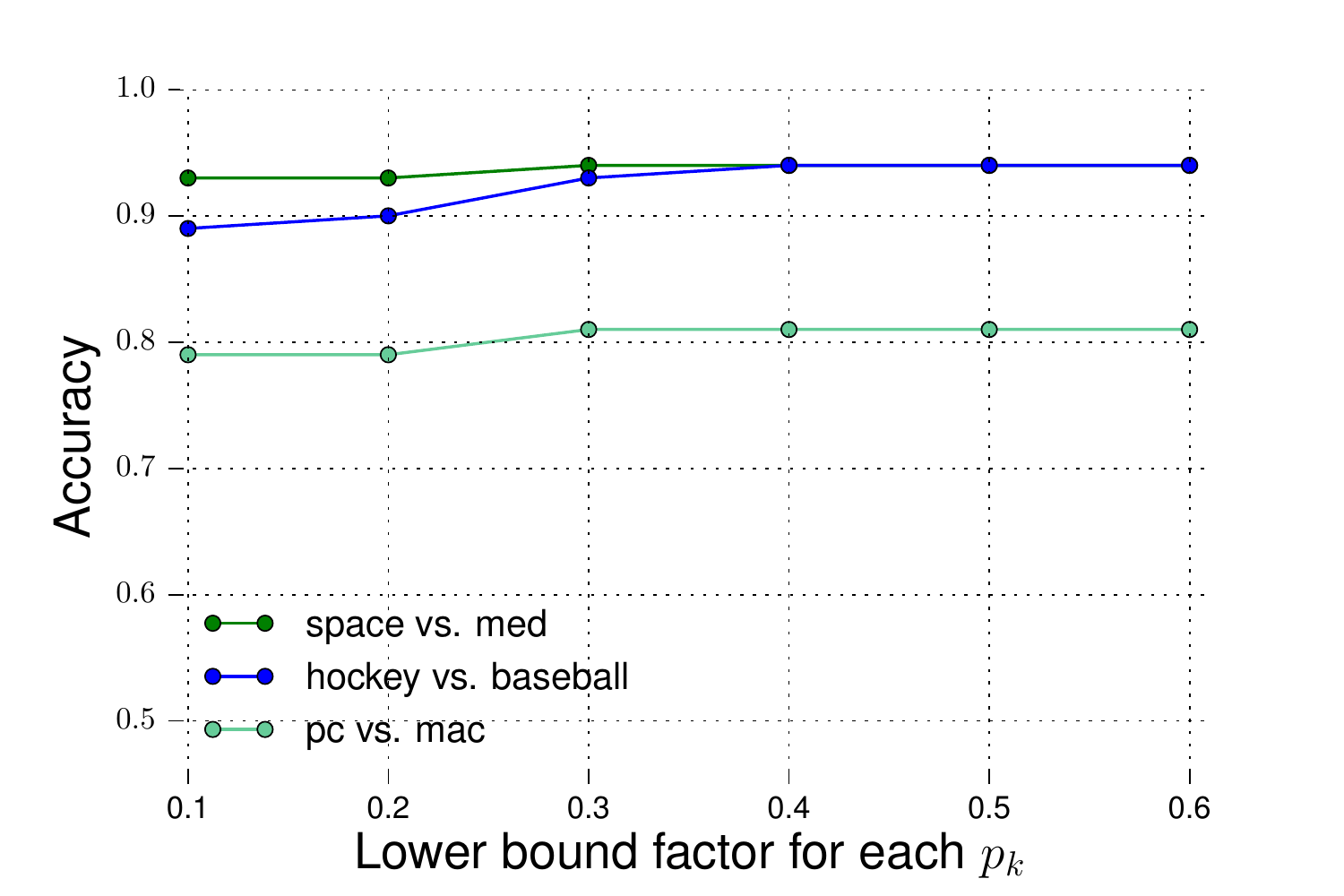}
 		% \caption{Caption 2}
 		\label{fig:subim2}
 	\end{subfigure}
 	\begin{subfigure}{1\textwidth}
    \centering
    (c)\\
 		\includegraphics[width=0.5\linewidth, height=4cm]{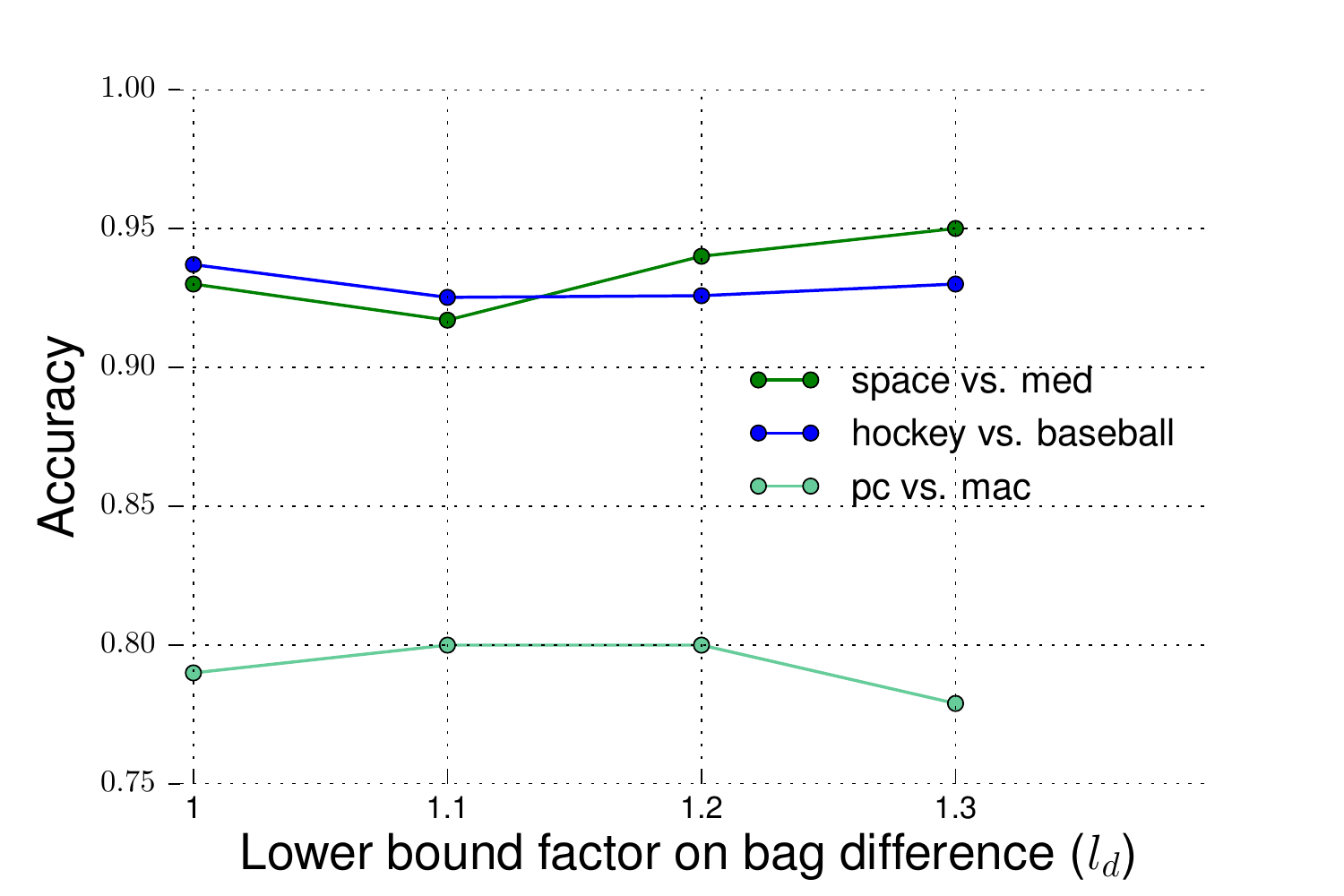}
 		% \caption{Caption 2}
 		\label{fig:subim0}
 	\end{subfigure}
% \begin{subfigure}{0.5\textwidth}\hspace{.4\textwidth}(d)\\
%  			\includegraphics[width=0.9\linewidth, height=4cm]{upper_diff}
%  			% \caption{Caption 2}
%  			\label{fig:subim3}
%  		\end{subfigure}
 	\caption{\small \textbf{Constraint effects}. We report accuracy results on a validation set. (a) Varying upper-bound factor on highest $p_k$. Performance remains stable for a long stretch until the upper bound on $p_{k_{max}}$ becomes too loose (b) Varying individual lower-bound factor (c) Varying lower bound on bag differences. Results remain fairly robust (with fluctuations due to small-sample noise). The graph stops abruptly at $l_d=1.3$ since beyond that point constraints are no longer feasible.}
 	\label{fig:20news}
 \end{figure*} }

Figure \ref{fig:20news} shows the results of our experiments.
In our initial setting, we take a fairly loose configuration of constraints to test our method's flexibility: $l_\text{d} = 1$ (no lower bound at all for bag differences), $l_\text{p} = 0.5$, and $u_m=1$. In each experiment we vary one factor, keeping the others fixed:
(a) upper bound on $p_{k_{max}}$, (b) individual lower bound, (c) lower bound on bag differences.

Notable in Figure \ref{fig:20news} is the overall robustness of the method to misspecified constraints. 
As $u_m$ is gradually increased, performance remains overall stable for a long stretch (\ref{fig:20news}a). However, when $u_m$ reaches extremely large values,  the upper bound on $p_{k_{max}}$ becomes too loose (reaching $1$) and robustness collapses. %The drop is more gradual for \textit{space} vs. \textit{med} than for the other two sets.
Increasing the lower bound on individual $p_k$ slightly improves results, by tightening constraints (\ref{fig:20news}b). Results remain fairly robust to overestimating the lower bound on bag differences by increasing $l_\text{d}$,with fluctuations due to small-sample noise (\ref{fig:20news}c). The graph stops abruptly at $l_d=1.3$ since beyond that point constraints are no longer feasible. 

Finally, we  compare results to the baselines of the previous Section. For our method, we fix $u_\text{m} = 1, l_\text{p} = 0.5, l_\text{d} = 1.33$ (with no upper bound on bag differences, as in previous sections). For the \textit{SVM using ``high vs. low"} baseline, we take bags $\mathcal{B}_1,\mathcal{B}_2$ as one class, and $\mathcal{B}_5,\mathcal{B}_6$ as the other (adding $\mathcal{B}_3,\mathcal{B}_4$ led to inferior results). Our method outperforms this naive baseline, and also competes with supervised SVM trained on considerable amounts of labeled examples (Table \ref*{table:20newsTable}). Given fewer labels, supervised SVM was inferior to our label-free method. 
%\vspace{-2pt}
\subsection{Simulation Study}
To further test the behavior of our algorithm, we conduct simulation studies on synthetic data. We use the built-in simulation function \textit{make\_classification} provided in python package scikit-learn \cite{Pedregosa2011sklearn} to generate data for a binary classification problem. We create three equally-sized bags of instances $\mathcal{B}_1, \mathcal{B}_2,\mathcal{B}_3$ for our training set, with label proportions $p_1,p_2,p_3$, respectively. We vary bag sizes $|\mathcal{B}_k|$ and proportions $p_k$, as well as the number of features (\textit{n\_features}), number of informative features (\textit{n\_informative}), and class separation (\textit{class\_sep}). %\textit{n\_clusters\_per\_class} was fixed at $2$. 

We apply our cross-validation procedure to select $\mathcal{C}$, using $3$ folds. We observe some typical behaviors, such as accuracy improvement with growing sample size. For instance, fixing $\textit{n\_features}=20$, $\textit{n\_informative} = 1$ and $p_1=0.4,p_2=0.3,p_3=0.2$, mean accuracy increases from $0.65$ with $|\mathcal{B}_k|=500$, to $0.77$ with $|\mathcal{B}_k|=1000$. 

Accuracy suffered with smaller gaps between bag proportions $p_k$. However, with increasing sample size our algorithm got better at handling minuscule differences between $p_k$. For example, fixing $p_1=0.4,p_2=0.35,p_3=0.33$, mean accuracy increases from $0.6$ with $|\mathcal{B}_k|=500$ to $0.65$ with $|\mathcal{B}_k|=1000$, and further increases to $0.67$ with $|\mathcal{B}_k|=1500$. 

Finally, we expect that labeled instances can improve performance, helping to counter bags that are very noisy or constraints that are not sufficiently tight. Preliminary experiments suggest that labeled instances can improve accuracy, but a comprehensive study of this effect is beyond the scope of this paper.

\begin{table*}[t]
	\centering
	\caption{\small \textbf{20 newsgroups results.} }
	\label{table:20newsTable}
	\begin{tabular}{|c|ccc|}
		\hline	
		{\bf Method} &{\bf med-space} &{\bf pc-mac} &{\bf baseball-hockey}  \\
		\hline
		{Bag constraints} & \textbf{0.94} & \textbf{0.81}  &  \textbf{0.94} \\
		``high vs. low" SVM & 0.82 &  0.62 &  0.64  \\ \hline
		Supervised SVM &  110 labels (0.93) & 95 labels (0.78)   &   140 labels (0.94)  \\	\hline
	\end{tabular}
\end{table*}

\section{Exploratory Analysis}
\label{sec:explore}

\begin{table*}[t]
	\centering
	\caption{\textbf{Geo-tagged tweets.} For each set of geographic constraints, we show some of the top positive and negative words resulting from running our method.}
    \label{table:tweets}
	\begin{tabular}{|p{2cm}p{3cm}|p{3cm}|p{3cm}|}
		\hline	
	{\bf Constraints}	&  &{\bf Positive terms} &{\bf Negative terms} \\
		\hline
        \begin{minipage}[t]{2cm}{\vspace{0cm} \includegraphics[width=2cm]{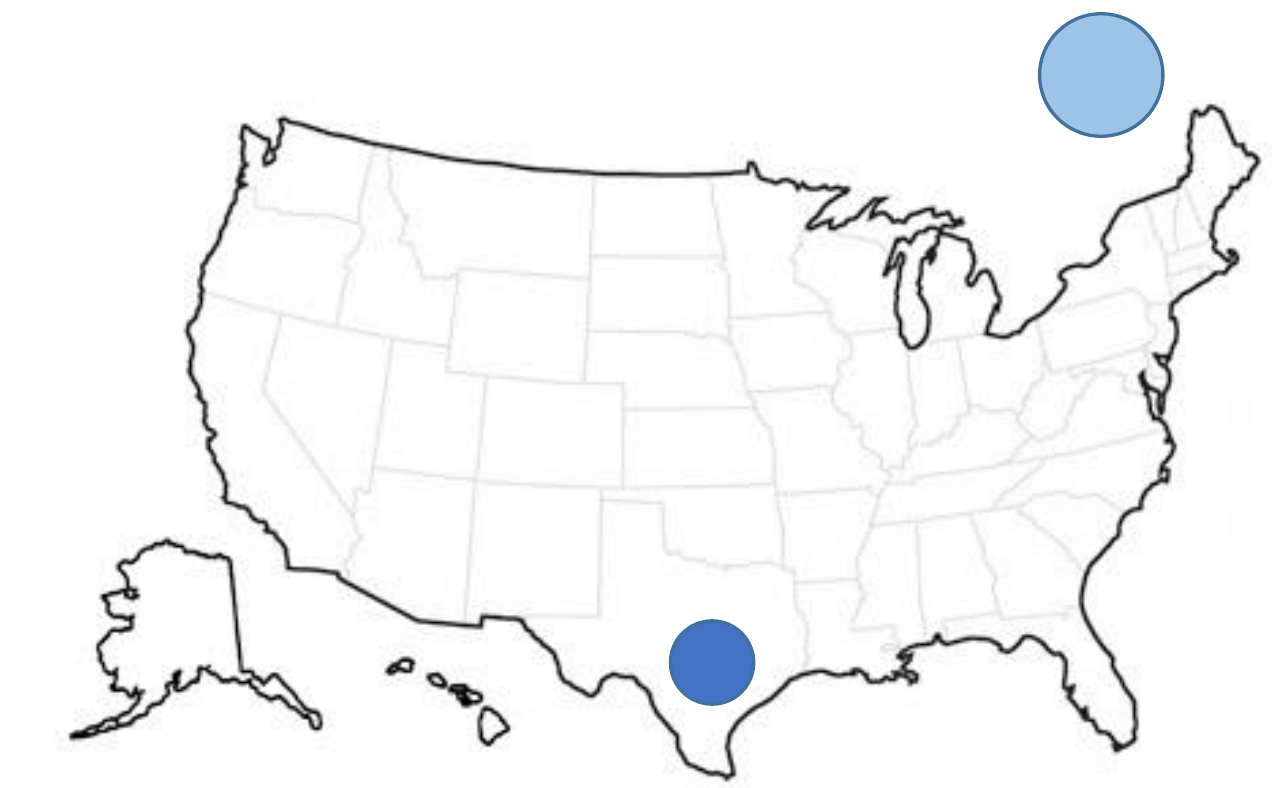}}  \end{minipage} & \textbf{French $>$ English:} Quebec $>$ Texas  & je, est, et, le, pour &   Houston, Texas, dallas, bro, tryna, boo
		\\ \hline
		\begin{minipage}[t]{2cm}{\  \\  \includegraphics[width=2cm]{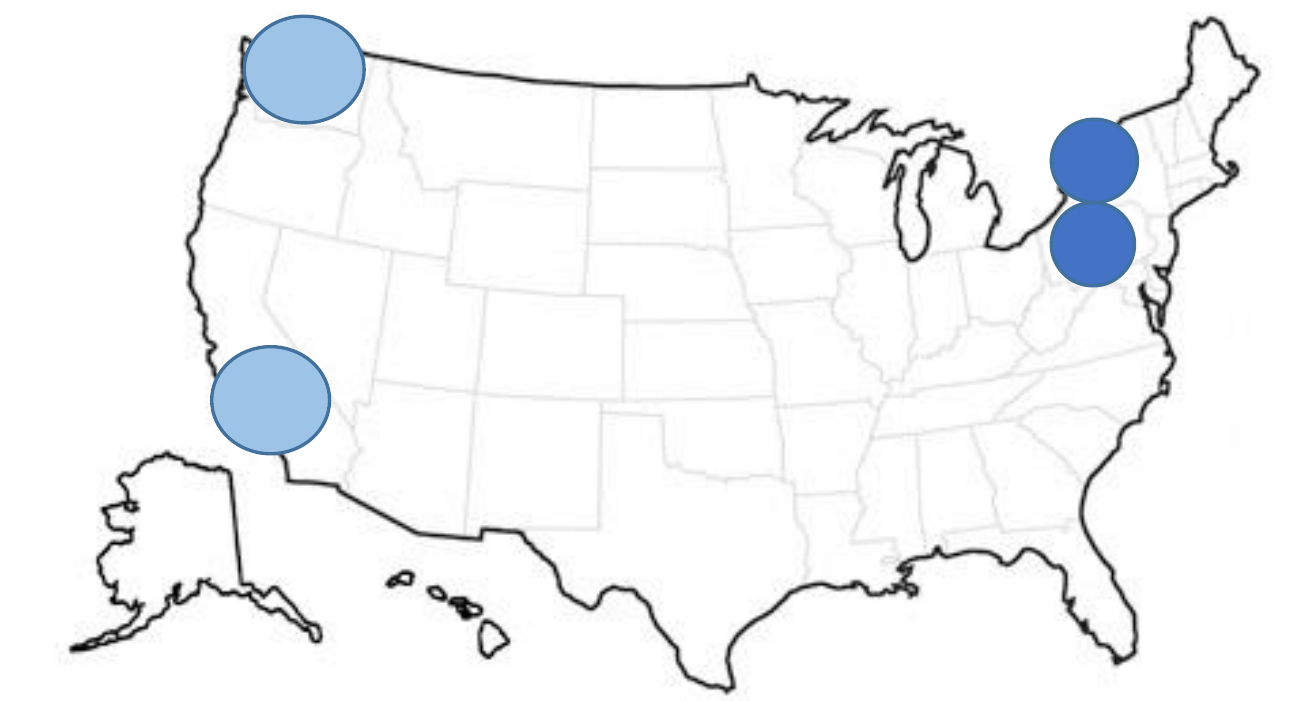}}  \end{minipage} & \textbf{East Coast $>$ West Coast:} CA $>$ NY, CA $>$ PA, WA $>$ NY, WA $>$ PA & hella, coo, fasho,  af, la, cali, san, washington & deadass, niggas, skool, wassup, dis, dat, philly, crib, lml, nah, dey, den \\ \hline
		\begin{minipage}[t]{2cm}{\  \\ \ \includegraphics[width=2cm]{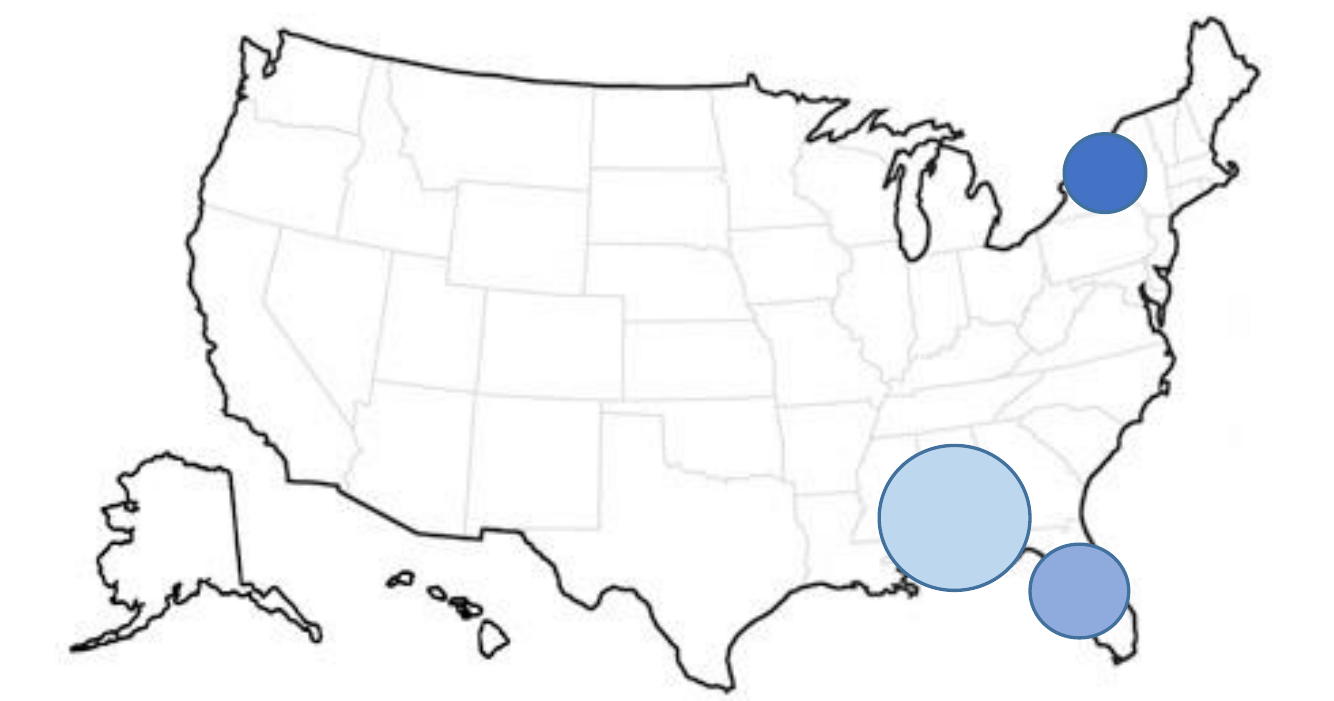}}  \end{minipage} & \textbf{Ranking by religiosity:} Alabama $>$ Florida $>$ New York & thank, easter, pray, road, trip, drove, loving, relationship, spring, folks, happy, dreams, laugh, friend & mad, bitches, neva, dis, dat, niggas,ova, spanish, girls, crazy, party, fun, high, dead \\
		\hline 
	\end{tabular}
\end{table*}

In previous sections we tackled classification problems with a clear objective. In this section our users have no specific classification in mind, but rather are interested in exploring the data. A sub-field within clustering allows users to guide the formation of clusters, usually in the form of pairwise constraints on instances (forcing data points to belong to the same cluster or to different clusters). A recent approach uses a maximum-margin framework \cite{Zhou13maxmargin}, which extends the supervised large margin theory (such as SVMs) to an unsupervised setting. Similarly, we adapt our method to the exploratory setting. Rather than using instance-level constraints on cluster membership, we use \emph{ranking} constraints based on prior knowledge -- or on hypotheses we would like to explore.

We used the {\bf Geo-tagged tweets} dataset, containing 377616 messages from 9475 geo-located microblog users over one week in March 2010 \cite{Eisen10Tweet}. The user base is likely dominantly composed of teens and young adults (as some of the examples below will make clear). We combine all tweets for each user, and reverse-geocode the GPS coordinates to obtain the corresponding state. 

%The authors explored the lexical variation between different regions, studying how language patterns vary across geographic contexts \cite{Eisen10Tweet}. They do so with geographic topic model, adding latent variables representing location to an overall generative process of language.

The dataset was used in \cite{Eisen10Tweet} to analyze regional dialects. The authors used a cascading topic model to model geographic topic variation. The observed output of the generative process includes the texts and GPS coordinates of each user.
%
%contains about $380,000$ microblog messages from about $9,500$ users in the United States, over one week in March 2010, with the user base likely dominantly composed of teens and young adults (as some of the examples below will make clear). These messages come with GPS coordinates (such as from mobile devices), enabling to assign geographic location to each user and message. In \cite{Eisen10Tweet} 
We pursue this line of exploration too, but rather than positing a generative model of language, we investigate how various constraints on differences between geographic locations interact with dialect.

%As mentioned in Section \ref{sec:tweets}, we explore how constraints on bags $\mathcal{B}$ representing states yield lexical variation. 
In Table \ref{table:tweets}, we show some of the constraints we explored and the resulting top positive and negative words. We start with a simple check with two bags  $\mathcal{B}_\text{Quebec} \succeq \mathcal{B}_\text{Texas}$, combining tweets from Quebec and Texas, respectively. We discover obvious differences in language, with strong positive weights corresponding to French words and negative weights to English. 
 
An ordinary classifier would likely recover similar results, as would standard unsupervised clustering algorithms.
However, our method allows to pursue richer, more \emph{expressive} constraints. First, we look into the difference between the East Coast and West Coast by imposing pairs of constraints such as $\mathcal{B}_\text{California} \succeq \mathcal{B}_\text{New York}$, $\mathcal{B}_\text{California} \succeq \mathcal{B}_\text{Pennsylvania}$. We recover various results previously highlighted by \cite{Eisen10Tweet}, such as the use of the slang terms ``fasho" (for sure) ``coo" (cool),``hella" in the West Coast, and ''deadass", ``wassup" and ``niggas" in the East Coast. Our results agree with findings by \cite{Eisen10Tweet,Eisen11socioling}, as well as suggest some potential new findings.
 
Finally, we look at a set of more expressive constraints, aiming to recover difference based on religiosity (or at least sociological confounders). We take states from the top, middle and bottom of a list of US states ranked by percentage of self-reported religiosity \footnote{\url{https://en.wikipedia.org/wiki/List_of_U.S._states_by_religiosity}}, and build sets of constraints that reflect this ordering. For instance, in Table \ref{table:tweets}, we show results for $\mathcal{B}_\text{Alabama} \succeq \mathcal{B}_\text{Florida} \succeq \mathcal{B}_\text{New York}$. 
Note that using such information in a standard classifier is unnatural. It is not clear how to construct the classes, and different splits could lead to very different results. Again, this artificial splitting is not required by our method.

We removed terms not in the wordnet \cite{wordnet98} lexicon to mitigate the effects of local vernacular and highlight deeper differences. The differences in language are quite striking. As we traverse from Alabama to Florida to New York, discourse shifts from words such as ``glad", ``loving", "happy", ``dreams", ``easter" and ``pray", to words including ``mad", ``bitches", ``crazy", ``party", ``fun", ``high" and other more profane content we spared from the reader. Similar results were obtained for other state tuples (e.g., Texas instead of Alabama).

Note that our method can be used for formulating new hypotheses. To test the hypotheses, more experiments (and often more data collection) are needed. We leave it up to sociologists to provide deeper interpretation of these results.
 
Our goal in this section was to use coarse prior information (in the form of relative rankings) for exploring a dataset. We note that the problem could be tackled with other approaches, such as topic models or classification. However, classification models assume a much stronger discriminative pattern or signal than taking a softer, weakly-supervised approach that seeks a direction (weight vector  $\mathbf{w}$) along which one bag of instances is ranked higher than another. While clustering with pairwise memberships constraints is well-studied, we demonstrate clustering with expressive pairwise \textit{ranking} constraints over sets. Many real-world settings naturally lend themselves to this formulation.

\section{Discussion and Criticism}
\label{sec:criticism}

One clear practical issue with our method is the source of the constraints. We have illustrated several real-world cases where it is plausible to attain rough constraints on label proportions within and between groups of instances. In previous work \cite{quadrianto2009estimating}, it is suggested that practitioners could sample from bags of instances to estimate label frequencies (e.g., in spam classification tasks). However, accurate estimations might require extensive sampling, exacting high costs. We thus relax this rather strong assumption, and propose that in many cases, it is possibly enough to get rough estimates. For example, after sampling $10$ instances, we might observe $9$ positives and only one negative, and rather conservatively declare ``$\mathcal{B}$ should have more than $50\%$ positives". This sort of statement could of course be made more rigorous with probabilistic considerations (e.g., confidence intervals). We have demonstrated that even with considerably mis-specified constraints, we are still able to achieve good performance across various domains. 

Furthermore, external sources of knowledge could be used to construct these constraints, such as previous surveys. In many cases taking exact figures from surveys (such as political polls) and expecting them to accurately reflect the distribution in new data is not realistic. This is the case, for instance, when looking at national political polls and wishing to extrapolate from them to new very different socio-demographic slices, such as Twitter users. Here too, we could use this external knowledge to \textit{approximately} guide our model, rather than dictate precise hard proportions the model should match.

\section{Related Work}
\label{sec:related}
There is a large body of work that is related to our problem.

\xhdr{Multiple Instance Learning}
The field of \emph{Multiple Instance Learning} (MIL) generally assumes instances come in ``bags", each associated with a label modeled as a function of latent instance-level labels, which can be seen as a form of weak supervision. MIL methods vary by the assumptions made on this function. For a comprehensive review of assumptions and applications, see \cite{Cheplygina14bags},\cite{Foulds10MIL}. Most work in MIL focuses on making bag-level predictions rather than for individual instances. Recently, \cite{Kotzias15deep} used a convolutional neural network to predict labels for sentences given document-level labels. 

\xhdr{Learning from Proportions}
A niche within MIL which has seen growing interest and is closely-related to this paper, is concerned with predicting instance-level labels from known label proportions given for each bag. \cite{quadrianto2009estimating} assume to be given bags of unlabeled examples, each bag with \textit{known} label proportions. Their method is based on estimating bag-means using given label proportions. The authors provide examples for scenarios in which such information could be available.
In \cite{Ruping10svmclassifier}, the authors represent each bag with its mean, and model the known class proportions based on this representative ``super-instance" with an SVM method, showing superior performance over \cite{quadrianto2009estimating}. In \cite{Felix13svm}, instance-level labels are explicitly modeled to overcome issues the authors raise with representing bags with their means, such as when data distribution has high variance. The fundamental property these and other approaches share is that bag proportions are assumed be known or easily estimated, an assumption we relax.

\xhdr{Classification with Weak Signals}
We applied our model to the problem of text classification when little or no labels are available but only a weaker signal. A vast amount of literature has tackled similar scenarios over the years, using tools from semi-supervised \cite{chapelle2006semi,Joachims99transduct}  active \cite{li2012multi,settles2010active,tong2002support} and unsupervised \cite{aggarwal2012survey} learning. Druck et al. \cite{Druck08features} apply generalized expectation feature-labeling (GE-FL) approaches, using ``labeled features'' given by an oracle that encode knowledge such as ``the word puck is a strong indicator of hockey''. In practice, a Latent Dirichlet Allocation (LDA) \cite{Blei03LDA} topic model is applied to the data to select top features per topic, for which a user provides labels. 
\cite{Settles11loop} propose a semi-supervised + active-learning method, with a human-in-the-loop who provides both feature-level and instance-level labels.  We are also able to use labeled instances to refine the learning process, allowing for a trade-off between the user's trust in the (typically few) labeled instances available, and prior knowledge on bag proportions.

Similar to the above work on learning from labeled proportions, \cite{wager2015clustering} considers a classification problem with no access to labels for individual training examples, but only average labels
over subpopulations. They frame the problem as weakly-supervised clustering. When using our method for exploratory analysis, it can also be seen as a weakly-supervised clustering algorithm, using information on partial ordering between bags rather than assuming known proportions, within a max-margin framework (somewhat akin to clustering using maximum-margin as in \cite{Zhou13maxmargin}). The seminal work of \cite{xing03side} uses side-information for clustering in the form of pairwise constraints on cluster membership (pairwise similarity). Much work has since been done along these lines. We incorporate pairwise constraints in our maximum-margin approach, though with pairs representing bags of instances, and partial ordering with respect to relative label proportions.

\xhdr{Robust Optimization}
Finally, robust optimization \cite{ben2009robust} research deals with uncertainty-affected optimization problems, by optimizing for the \emph{worst-case} value of parameters. Because of its worst-case design, robust optimization can do poorly when the constraints are not tight. Our method, on the other hand, is designed to handle rough estimates and loose constraints.

%\begin{figure}
%\centering
%\includegraphics{fly}
%\caption{A sample black and white graphic.}
%\end{figure}

% \begin{figure}
% 	\centering
% 	\includegraphics[height=3in, width=5in]{negative_step_2}
% 	\caption{Negative step 2}
% \end{figure}

\section{Conclusions and Future Work}
\label{sec:conc}
%While we believe examples given in above sections reflect real-world scenarios, they are admittedly somewhat artificial. We thus plan to test our newly proposed machine-learning setting and approaches on real cases arising in practice, testing whether our assumptions hold. In particular, we would like to test the assumption of stability -- whether our preliminary results that show good robustness to mis-specified constraints, carry on to other, more difficult cases. 

%Our goal in this paper was to propose a new scenario, formalize the problem and provide simple solutions. We have empirically studied our algorithm and suggested some intuitive insights. Theoretically studying the behavior of our algorithm under different settings was outside the scope of our current work. It is of interest to analytically understand, for instance, how constraint tightness affects performance on unseen data, obtain convergence guarantees, and provide generalization error bounds. This, in turn, could perhaps lead to better algorithms with theoretical justifications. 
%\dnote{merge above into}

In this paper we proposed a new learning setting where we have bags of unlabeled instances with loose constraints on  label proportions and difference between bags. Thus, we relax the unrealistic assumption of known bag proportions. 

We formalized the problem as a bi-convex optimization problem and proposed an efficient algorithm. We showed how, surprisingly, our classifier performs well using very little input. We also demonstrated how the algorithm can guide exploratory classifications. 

%One direction for future work is to use better representations than the bag mean in feature-space. One way to do so, for instance, could be to take the soft-maximum of a bag, leading to convex-concave constraints \cite{Hu08ranking} and a heavier optimization problem. Another possible approach, would be to iteratively ``prune" bags of low-scoring examples (with a low $\mathbf{w}^{T}\varphi(\mathbf{x}_i)$), or assign ``importance weights" to examples in each bag, inversely proportional to their score such that strongly positive examples would dominate. Another direction could be to follow \cite{Felix13svm} and directly model the unobserved labels.
We have empirically studied the behavior of our algorithm under different types of constraints. One direction for future work is to analytically understand, for instance, how constraint tightness affects performance, obtain convergence guarantees, and provide generalization error bounds. This, in turn, could perhaps lead to better algorithms with theoretical justifications. 

Finally, the relative-proportions setting is very natural, and can be found in various domains. We believe that this line of work will have interesting implications regarding privacy and  anonymization of data -- in particular, the amount of information one can recover using only weak, aggregated signals. 
%\dnote{spellcheck the whole thing}

\xhdr{Acknowledgments} The authors thank the anonymous reviewers and Ami Wiesel for their helpful comments. Dafna Shahaf is a Harry\&Abe Sherman assistant professor, and is supported by ISF grant 1764/15 and Alon grant.
%\end{document}  % This is where a 'short' article might terminate

\bibliographystyle{abbrv}
\bibliography{sigproc}  % sigproc.bib is the name of the Bibliography in this case
% ACM needs 'a single self-contained file'!
%

\end{document}